\documentclass{article}

\usepackage[eandd, preprint]{neurips_2026}
\usepackage{graphicx}

\usepackage[utf8]{inputenc} %
\usepackage[T1]{fontenc}    %
\usepackage{hyperref}       %
\usepackage{url}            %
\usepackage{booktabs}       %
\usepackage{amsfonts}       %
\usepackage{nicefrac}       %
\usepackage{microtype}      %
\usepackage{xcolor}         %
\newcommand{\locus}{\textsc{locus}}
\usepackage{booktabs}
\usepackage{tabularx}
\usepackage{array}
\usepackage{tikz}
\usetikzlibrary{positioning, arrows.meta, calc}

\usepackage[most]{tcolorbox}
\newtcblisting{promptbox}[1]{
  enhanced, breakable,
  colback=gray!4, colframe=black!55,
  fonttitle=\bfseries, coltitle=white,
  title=#1,
  listing only,
  listing options={
    basicstyle=\small\ttfamily,
    breaklines=true,
    breakatwhitespace=true,
    columns=fullflexible,
    keepspaces=true,
  },
  left=6pt, right=6pt, top=4pt, bottom=4pt,
  boxrule=0.5pt, arc=2pt,
}

\title{Freeing the Law with LOCUS: \\ A Local Ordinance Corpus for the United States}

\author{%
  \textbf{Denis Peskoff}$^{*1,2}$ \quad
  \textbf{Joe Barrow}$^{*3}$ \quad
  \textbf{Christopher Vu}$^{1}$ \quad
  \textbf{Diag Davenport}$^{1,2}$ \\[2pt]
  $^{*}$Equal contribution \quad
  $^{1}$UC Berkeley \quad
  $^{2}$School of Information \quad
  $^{3}$Independent \\
  \texttt{\{dpeskoff, diag\}@berkeley.edu}
}
\begin{document}

\maketitle

\begin{abstract}
Progress in legal AI increasingly depends on access to authoritative legal text at scale. Yet one of the most consequential layers of American law remains largely absent from existing machine-readable corpora: local ordinances. Local codes govern zoning, housing, business licensing, public health, noise, animal control, and many other domains of everyday regulation, but they are fragmented across vendor platforms designed for human browsing rather than bulk research access. We introduce \locus{}—the \textbf{L}ocal \textbf{O}rdinance \textbf{C}orpus for the \textbf{U}nited \textbf{S}tates—a comprehensive corpus and county-harmonized access layer for U.S. municipal and county ordinance codes. The raw corpus, available for release to researchers, represents nearly all publicly available municipal and county ordinance codes. The resulting raw corpus contains codes from 9,239 cities and counties. A smaller county-harmonized LOCUS access layer provides coverage for the largest 2,309 of 3,144 U.S. counties, accounting for a majority of the population. 
We use OCR to handle the myriad of document formats that have kept the law from being a public resource. We release the corpus with coverage metadata to support reproducibility, downstream legal AI research, and the incremental expansion of machine-readable access to local law. We train a collection of ModernBERT-based classifiers and scorers to facilitate analyzing U.S. local law among several dimensions, such as opacity and paternalism, that have not previously been studied at this scale. \locus{}-v1 and its derivative models are available at: \href{https://huggingface.co/datasets/LocalLaws/LOCUS-v1}{https://huggingface.co/datasets/LocalLaws/LOCUS-v1}
\end{abstract}

\begin{figure}[h!]
    \centering
    \includegraphics[width=.9\linewidth]{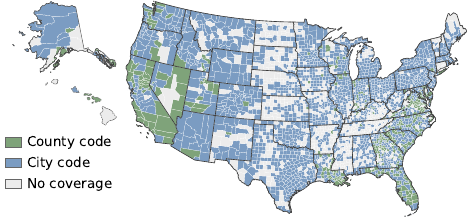}
    \caption{\locus{} represents the longest digitally available code---city or county---for each county.}
    \label{fig:coveragemap}
\end{figure}

\section{What it means to "free the law"}

\begin{figure}[t]
  \centering
  \includegraphics[width=0.5\linewidth]{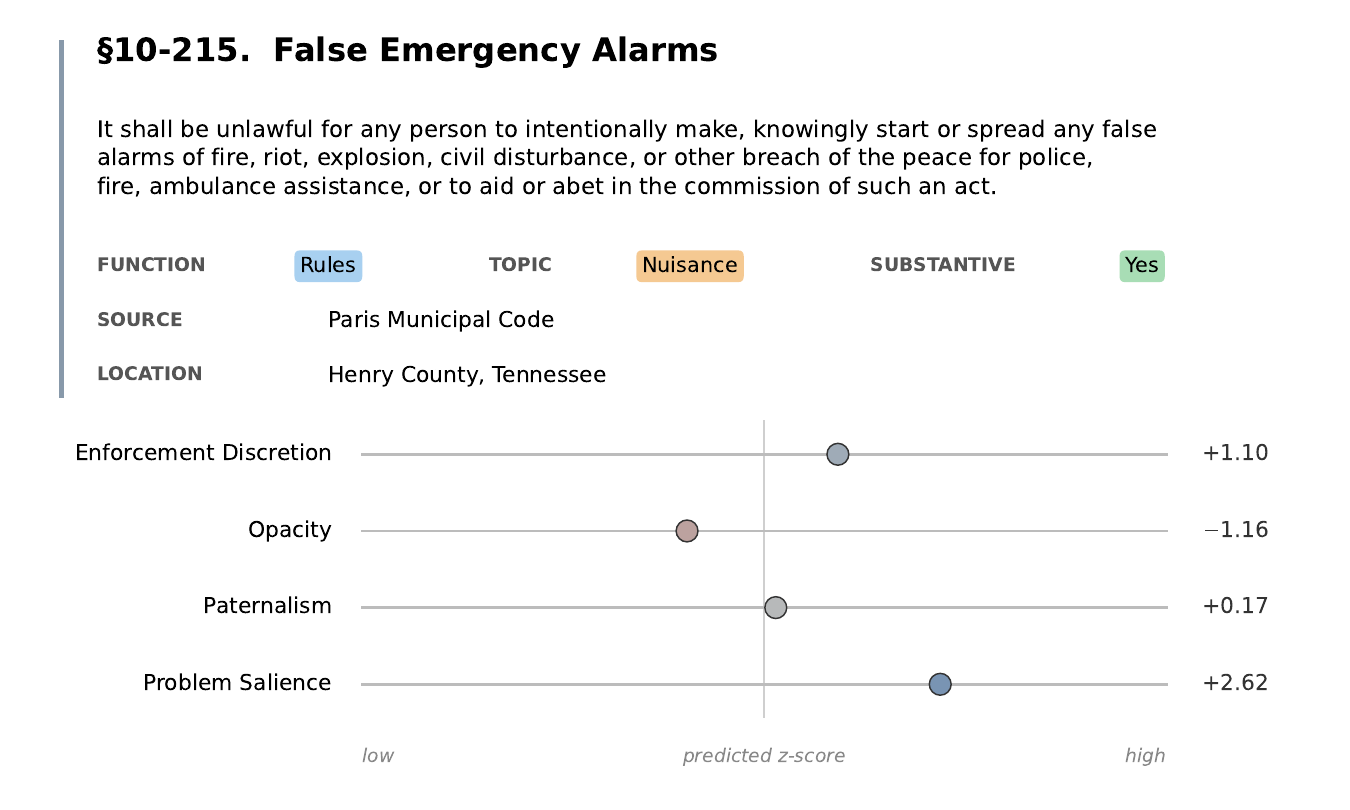}\hfill
  \includegraphics[width=0.5\linewidth]{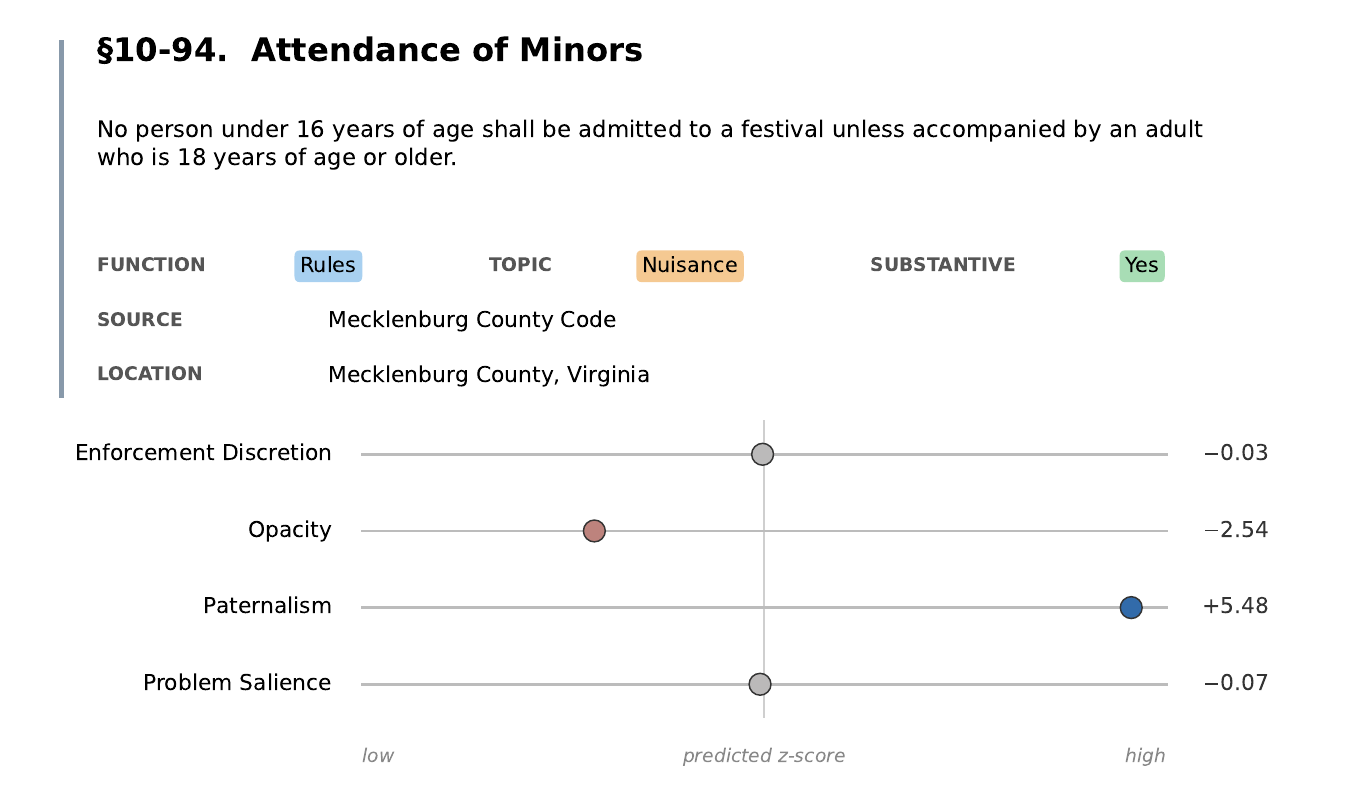}
  \caption{Two example ordinances with predicted scores (in standard units) on four axes (\textit{opacity, enforcement discretion, paternalism}, and \textit{salience.}) produced by ModernBERT regressors (\S\ref{sec:evaluations}) and function/topic labels produced by ModernBERT classifiers (\S\ref{subsec:annotation}). Together they demonstrate the per-ordinance analysis enabled by \locus{}. }
  \label{fig:axes-demo}
\end{figure}

Legal AI systems increasingly operate over statutes, cases, regulations, contracts, and administrative materials \citep{chalkidis2022lexglue,henderson2022pile,guha2023legalbench}. 
This expansion has been accompanied by domain-specific resources for case law \citep{zheng2021casehold}, contracts \citep{hendrycks2021cuad,koreeda2021contractnli,tuggener2020ledgar}, and statutory reasoning \citep{holzenberger2020sara}.
Despite this, they still lack systematic access to one of the most consequential layers of American law: local ordinances. These codes govern zoning, housing, building permits, business licensing, public health, noise, signs, animal control, and other domains of everyday regulation. For many questions faced by residents, businesses, landlords, and local governments, the relevant legal text is not only federal or state law, but a municipal or county code.

Local law is not merely another collection of statutes. It is a layered system of legal authority. State statutes, county ordinances, municipal codes, home-rule provisions, charters, preemption doctrines, and issue-specific delegations can all interact. Whether a state rule, county rule, or municipal rule controls is often not obvious in the abstract and may depend on the legal domain. This makes local law a particularly important setting for legal AI: a useful system must not only retrieve text, but identify the relevant jurisdictional layer and reason about overlap, delegation, and conflict among sources of authority.

We introduce \textbf{\locus{}-v1}, a large-scale corpus and county-harmonized access layer for U.S. local ordinances. The first release of \locus{} adopts a deliberately transparent simplification: for each U.S. county, we record the most substantial available local code among the county ordinance code and the ordinance code of the county's largest municipality, using document length as a reproducible proxy for local-law coverage. This representation does not purport to decide which local authority controls every legal question. Rather, it provides a common geographic substrate on which local legal text can be searched, compared, and connected to population, geographic, Census, and policy data.

The need for such a dataset arises because local law is public but not practically available as a national research corpus. \citet{GeorgetownLawLibrary2026}, the most applied-to law school in the United States comments, ``\textit{there is unfortunately no single source where you can find a comprehensive collection of all municipal codes.}''
U.S. local codes are fragmented across commercial vendor platforms designed for in-browser reading rather than bulk research access. Vendors expose different navigation structures, print workflows, dynamically generated PDFs, and jurisdiction indexes. No central registry maps every county or municipality to its hosting platform, and no vendor provides a complete machine-readable index of all jurisdictions it hosts. As a result, constructing a national corpus requires discovering where each code lives, extracting it through platform-specific workflows, validating the resulting artifacts, and harmonizing them to a common unit of analysis.

We leave full issue-specific hierarchy and conflict modeling to later releases and benchmark tasks.
This staged design reflects both the legal complexity of determining controlling authority and the need to preserve uncontaminated evaluation settings for future legal-reasoning benchmarks.

\locus{} enables a new class of legal AI and empirical legal studies applications. At the retrieval layer, it supports search and question answering over local rules whose terminology varies substantially across jurisdictions. At the representation layer, it enables structured extraction of regulated activities, permits, fees, penalties, effective dates, and cross-references. At the reasoning layer, it creates a foundation for benchmarks that test whether systems can navigate multiple layers of law, identify the relevant jurisdictional authority, and reason about state-local or county-municipal overlap. By making local law observable at national scale, \locus{} turns a fragmented body of public legal authority into infrastructure for legal retrieval, regulatory extraction, comparative policy analysis, and legal-domain language model evaluation.

We provide a summary of our corpus (§\ref{sec:properties}), decision points necessary to create it (§\ref{sec:construction}), evaluations of the corpus (§\ref{sec:evaluations}), and a discussion of how this can improve our understanding of the legal system (§~\ref{sec:discussion}).

\section{Related Work}
Studying the law has been important in society for centuries~\citep{holmes1897path}.  In the Information Age, the law has become both immediately accessible but increasingly complicated.  We are not the first to create corpora for legal NLP~\citep{steinberger2006jrc,aletras2016predicting,livermore2017supreme,caselaw_access_project}.  
Neural network era corpora such as ECHR~\citep{chalkidis2019neural} and pile of law~\citep{henderson2022pile} contain case law, court and administrative opinions, and legal codes but not the local law.  
The 162 tasks in LegalBench~\citep{guha2023legalbench} draw heavily from contracts and merger agreements and none involve local ordinances.  

Access to the law is a historical challenge which has been reshaped in part by the internet.  
Georgia v. Public.Resource.Org, Inc., No. 18-1150 (decided April 27, 2020)
~\citep{georgia_v_public_resource_2020} upheld that laws, statutes, and court decisions are public domain, in so far as digital content goes.
Since that time the rise of large language models and other modern techniques has enabled intelligent data processing on an unprecedented scale; standardizing over 9,239 one-thousand page documents would not have been feasible several years ago.  
Local laws have been understudied in part due to data access that we hope \locus{} will resolve.

\section{Properties of LOCUS}
\label{sec:properties}

\begin{figure}
    \centering    \includegraphics[width=\linewidth]{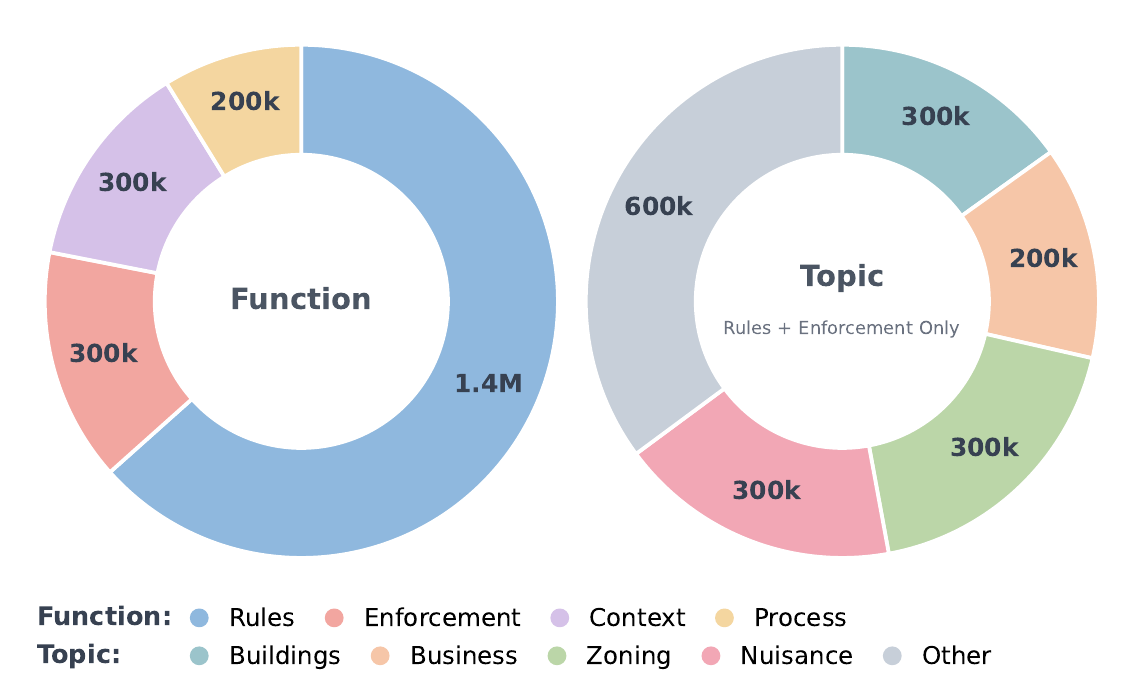}
    \caption{We annotate our corpus at the chunk level along its \textit{Function}, and the substantive laws \{Rules and Enforcement\} according to the \textit{Topic} referenced. Table~\ref{tab:locus-examples} provides example texts.}
    \label{fig:pies}
\end{figure}

\begin{table}[t!]
\centering
\setlength{\tabcolsep}{5pt}
\renewcommand{\arraystretch}{1.15}
\begin{tabularx}{\columnwidth}{@{}>{\raggedright\arraybackslash}p{0.15\columnwidth} X@{}}
\toprule
\textbf{Label} & \textbf{Representative example} \\
\midrule
\multicolumn{2}{@{}l}{\textit{Function}} \\
\cmidrule(lr){1-2}
\textsc{Rules}       & No direct seller shall engage in direct sales within the city without receiving a permit for that purpose as provided herein. \\
\textsc{Enforcement} & Any Code Enforcement Officer may issue notices of violation and administrative citations, inspect public and private property, and enforce any available administrative remedies. \\
\textsc{Structural}  & Park Regulations 973.01 Adoption and purpose. 973.02 Powers. 973.03 Enforcement. 973.04 Application to concessions. \\
\textsc{Context}     & The purpose of this notice and review procedure is to notify the public of the permit review process for development proposed in areas having identified significant resources and functional values. \\
\textsc{Process}     & Application for a license required by this division shall be filed with the city clerk on forms provided for that purpose. \\
\addlinespace[3pt]
\multicolumn{2}{@{}l}{\textit{Topic}} \\
\cmidrule(lr){1-2}
\textsc{Buildings}   & The registered design professional shall submit sufficient technical data to substantiate the proposed alternative engineered design and prove that the performance meets the intent of this code. \\
\textsc{Business}    & No direct seller shall engage in direct sales within the city without receiving a permit for that purpose as provided herein. \\
\textsc{Zoning}      & Where a change of use of an existing structure requires additional parking or other requirements applicable to a new use, a site plan shall be submitted for review. \\
\textsc{Nuisance}    & Bike paths may be used only for the operation of bicycles and pedestrian use. \\
\textsc{Other}       & Monies appropriated for salaries, wages and related benefits shall not be used for general operations, capital outlay, or other purposes without recommendation from the Mayor and specific approval of a majority of the council. \\
\bottomrule
\end{tabularx}
\caption{Representative examples for the five \textit{Function} labels and the five merged \textit{Topic} labels in \textsc{Locus}. All items in the topic group are annotated as \textit{Rules} or \textit{Enforcement} in their function.}
\label{tab:locus-examples}
\end{table}

Our corpus benefits both the technical and social science communities by providing valuable data and insight.  We discuss the harmonized LOCUS access layer and additional data provided for researchers.

\subsection{A County-Harmonized Access Layer}

\locus{} adopts a transparent simplification: for each U.S. county, it identifies the most substantial available local code among the county ordinance code and the ordinance code of the county's largest municipality. This design does not purport to determine which layer of law controls in every doctrinal context. Instead, it provides a reproducible substrate for retrieval, comparison, and future benchmarks on state–county–municipal legal reasoning.

 Figure~\ref{fig:pies} summarizes our publicly released corpus: 2,211,516 chunks of text, out of which the majority are judged to be substantive laws in nature.  We define substantive as concerned with rules or enforcement, rather than any text that is purely structural, process-oriented, or purely context; the majority of our annotations are rules. These substantive laws deal with four major categories: buildings, business licensing,  zoning, and nuisance.  
 The remainder, roughly a third of the laws are categorized with near 90\% precision as other.  We investigate the headers of these chunks and find that other constitutes topics such as government, employment matters, and animal regulation (this last category makes Alaska have a disproportionately large share of 'other' chunks).  Table~\ref{tab:locus-examples} provides examples that illustrate the diversity of laws.  

\subsection{Additional Data for Researchers}
In addition to the released data, we collect an additional 7,000 documents of other cities and counties.
We intend to make this data available to researchers with signed release similar to MIMIC~\citep{johnson2016mimic,johnson2023mimic}.
Given current LLM ingestion policies, we believe this is necessary for any future evaluation of local law coverage by foundational models~\citep{dahl2024hallucinating}.

\section{Constructing \locus{}}
\label{sec:construction}

An overview of the pipeline is shown in Figure~\ref{fig:pipeline}.

\definecolor{stageFill}{HTML}{F5F5F3}
\definecolor{stageEdge}{HTML}{C9C9C5}
\definecolor{stageText}{HTML}{1F2937}

\definecolor{classifyFill}{HTML}{FBEEDA}
\definecolor{classifyEdge}{HTML}{D9B888}
\definecolor{classifyAccent}{HTML}{8E4F1C}

\definecolor{scoreFill}{HTML}{E3EDE7}
\definecolor{scoreEdge}{HTML}{8DB1A4}
\definecolor{scoreAccent}{HTML}{1F5E54}

\definecolor{arrowColor}{HTML}{4B5563}

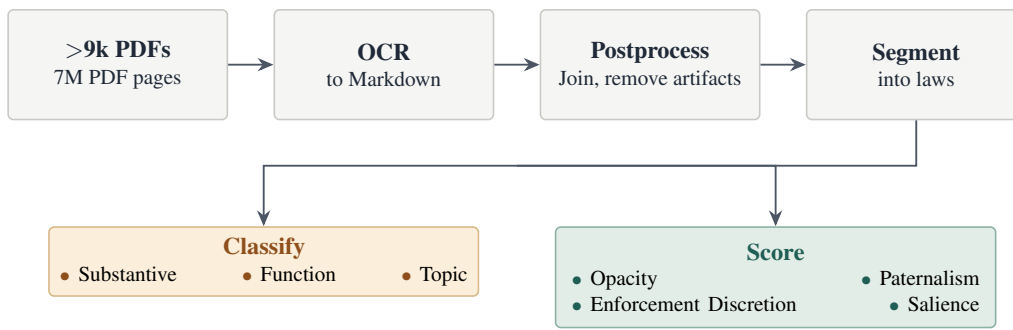
\begin{figure*}[t]
\centering
\begin{tikzpicture}[
    font=\small,
    >=Stealth,
    stage/.style={
        draw=stageEdge,
        rounded corners=2pt,
        minimum height=1.45cm,
        minimum width=2.9cm,
        align=center,
        line width=0.6pt,
        fill=stageFill,
        text=stageText,
        inner sep=4pt
    },
    classifyBox/.style={
        draw=classifyEdge,
        rounded corners=2pt,
        line width=0.6pt,
        fill=classifyFill,
        text width=5.4cm,
        align=center,
        inner sep=4pt
    },
    scoreBox/.style={
        draw=scoreEdge,
        rounded corners=2pt,
        line width=0.6pt,
        fill=scoreFill,
        text width=5.4cm,
        align=center,
        inner sep=6=4pt
    },
    arrow/.style={
        -{Stealth[length=6pt, width=5pt]},
        line width=0.7pt,
        draw=arrowColor
    },
    line/.style={
        line width=0.7pt,
        draw=arrowColor
    }
]

\node[stage] (pdfs) {%
    \textbf{$>$9k PDFs}\\
    \footnotesize 7M PDF pages
};
\node[stage, right=0.6cm of pdfs] (ocr) {%
    \textbf{OCR}\\
    \footnotesize to Markdown
};
\node[stage, right=0.6cm of ocr] (post) {%
    \textbf{Postprocess}\\
    \footnotesize Join, remove artifacts
};
\node[stage, right=0.6cm of post] (segment) {%
    \textbf{Segment}\\
    \footnotesize into laws
};

\draw[arrow] (pdfs) -- (ocr);
\draw[arrow] (ocr) -- (post);
\draw[arrow] (post) -- (segment);

\coordinate (pmid) at ($(pdfs)!0.5!(segment)$);
\coordinate (pmid_bot) at (pmid |- pdfs.south);

\node[classifyBox, anchor=north east] at ($(pmid_bot) + (-0.5, -1.4)$) (classify) {%
    {\bfseries\textcolor{classifyAccent}{Classify}}\\
    \footnotesize
    {\scriptsize\textcolor{classifyAccent}{$\bullet$}}~Substantive \hfill
    {\scriptsize\textcolor{classifyAccent}{$\bullet$}}~Function \hfill
    {\scriptsize\textcolor{classifyAccent}{$\bullet$}}~Topic
};

\node[scoreBox, anchor=north west] at ($(pmid_bot) + (0.5, -1.4)$) (score) {%
    {\bfseries\textcolor{scoreAccent}{Score}}\\
    {\footnotesize
    {\scriptsize\textcolor{scoreAccent}{$\bullet$}}~Opacity \hfill {\scriptsize\textcolor{scoreAccent}{$\bullet$}}~Paternalism\\
    {\scriptsize\textcolor{scoreAccent}{$\bullet$}}~Enforcement Discretion \hfill {\scriptsize\textcolor{scoreAccent}{$\bullet$}}~Salience}%
};

\coordinate (junction) at ($(pmid_bot) + (0, -0.6)$);

\draw[line] (segment.south) |- (junction);
\draw[arrow] (junction) -| (classify.north);
\draw[arrow] (junction) -| (score.north);

\end{tikzpicture}
\caption{%
    \textbf{Processing pipeline.}
    A corpus of more than 9{,}000 PDFs (7M pages total) is OCR'd into Markdown,
    cleaned, and segmented into individual laws. Each segment is then
    independently \emph{classified} for function, topic, and substance and \emph{scored}
    on four normative dimensions.%
}
\label{fig:pipeline}
\end{figure*}

\subsection{Collecting the data}
The original raw corpus contains 9,239 valid PDFs totaling approximately $\sim$80~GB. 
Constructing \locus{} required solving a coupled systems and legal-data problem across thousands of jurisdictions.
Our pipeline uses browser automation and vendor-specific download logic to collect municipal and county codes from major hosting platforms. 
The construction process surfaced several nontrivial failure modes, including server-side PDF assembly limits, filename collisions among non-unique municipality names, hidden interface thresholds, 15 second crawl delays, anti-bot measures, and multi-county consolidated cities. 
Addressing these failures required targeted recovery techniques rather than a single generic scraper.
Furthermore, we manually collect self-hosted or pdf-restricted codes for cities and counties which are not covered by this methodology.

\subsection{Identifying salient laws}
\label{identification}

Given the huge amount of data, and the diversity of its content and format, we employ a two-level zero-shot approach as the initial labeling approach.  Given that our data is being ingested in thousands of different formats after OCR, we need to remove structural content (i.e., stray headers, table of contents) and identify the substantive chunks.

After preliminary investigation of Anthropic and Gemini, we settle on OpenAI's GPT-5.4 as a fast and reliable annotator for this data~\citep{openai2026gpt54}.  After comparing a 500 sample of 5.4 mini and nano, we select nano as a cost-effective and only marginally worse option for large-scale annotation.  Inspired by LLM-as-a-Judge~\citep{zheng2023judging}, we evaluate the 5.5\% of annotations deemed most challenging with a much more expensive GPT-5.4 model.  The model agrees on 64,977 out of 108,889  predictions. The more advanced model often decreases its predictions of rules in favor of process and enforcement.  Crucially, no models hesitated in identifying structural content, which was ultimately removed from our release.  We intend to maintain this dataset and hope to get support from the LLM and law communities in improving these annotations as we update the corpus.  Ideally, direct evaluation by lawyers and judges would enable us to exceed the limitations of LLM-as-a-Judge.
 
\subsection{OCR and Processing}
\label{ocr}

The ordinances are stored in diverse layouts and formats, including single- and double-column layouts, born-digital, exported, and scanned documents, etc.
To best handle this diversity, the pipeline for building LOCUS starts by running optical character recognition (OCR) to convert every image of a page to Markdown.

We accomplish this with LightOnOCR-2-1B~\citep{taghadouini2026lightonocr}, an open 1B parameter vision-language model (VLM) based on Qwen-3~\citep{bai2025qwen3} finetuned on 16MM PDF pages that scores highly on a standard OCR benchmark, OlmOCR-Bench~\citep{poznanski2025olmocr}.
LightOnOCR-2-1B generates Markdown text from a page image.
We find that this model is robust to the diversity of the raw ordinances, consistently generating correct text in natural reading order.

The rest of our post-processing pipeline consumes the unified Markdown output to stitch together laws across pages.
We strip artifacts such as repeated headers, footers, and page numbers, and merge content that crosses pages such as paragraphs and tables.
The next stage of this post-processing pipeline is to segment the joined content into individual laws, identifying section and subsection headers.

The final step of our post-processing pipeline is to classify the substantivity, function, and topic of each extracted law.
We discuss the construction of these classifiers in \ref{subsec:annotation}.
Each is trained on the roughly 100M parameter ModernBERT-base~\citep{warner2025smarter} encoder, which enables us to efficiently run inference on every law.
Segments that are classified as purely structural, rather than containing any laws, are omitted from the dataset.

The raw ordinances are contained in roughly 7M pages.
We are able to scale our OCR pipeline on Modal\footnote{\href{https://modal.com}{https://modal.com}}.
Given the relatively small size of LightOnOCR-2-1B and Modal's batch inference support, we were able to efficiently run the entire pipeline and process documents across all formats at roughly \$0.30 per 1,000 pages.

\subsection{Annotating the Law}
\label{subsec:annotation}

To organize the ordinances, we develop three classifiers: \textbf{substantivity}, \textbf{function}, and \textbf{topic}.
A breakdown of the label space and selected examples are shown in Table~\ref{tab:locus-examples}.

We build these classifiers by sampling 100{,}000 laws from the pipeline discussed in the previous section and using GPT-5.4-nano to annotate each of them.
The resulting labels are used to train a ModernBERT classifier~\citep{warner2025smarter}, which can be efficiently used for inference across the rest of the dataset.
The classifiers are trained using 80{,}000 samples for training, 10{,}000 for parameter sweeps, and finally evaluated on a 10{,}000 instance subset.

From this collection, \locus{}-v1 derives a county-harmonized release that records a representative local-law artifact for each covered county, together with the structured metadata from the classifiers.

\subsection{Creating a Harmonized Access Layer}

Our access layer illustrated in Figure~\ref{fig:coveragemap} is built by a simple algorithm run on all the codes: for every county in the United States, is there an existing county code and an existing city code, ideally from the largest city in that county? If both exist, pick the longest by page length.  This is an imperfect process but length of code and population of jurisdiction were correlated.\footnote{Counties run on average slightly shorter than cities, but we opted for an easily interpretable selection algorithm rather than introducing weights; this did not dramatically impact the final selection as certain states, such as Maryland, have much more powerful counties than cities.}  By doing this, we are able to provide a code for counties \textit{representing} 94\% of the United States by population.  Since for example the second order city or the population living in the county outside the city are not captured by this, this access layer applies to a smaller literal population than the full data.  

\section{A Dimensional Analysis of Local Laws}
\label{sec:evaluations}

\begin{figure}[t!]
    \centering
    \includegraphics[width=\linewidth]{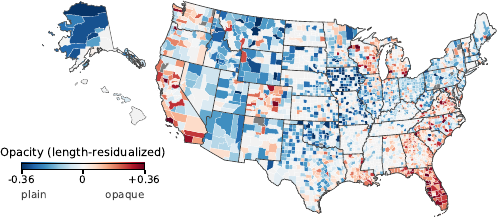}\\
    \includegraphics[width=\linewidth]{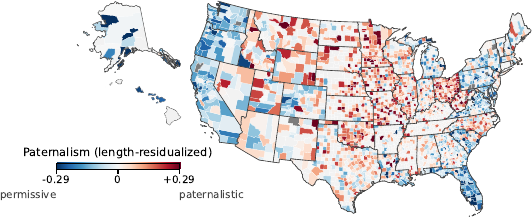}
    \caption{The opacity and paternalism of laws varies across the country.  \locus{} facilitates studying the laws for macro trends such as discovering that Florida law is opaque but not paternalistic.}
    \label{fig:distribution}
\end{figure}

By linking the text of the laws to the locales in which they apply, \locus{}-v1 opens the door for new types of analysis.
In addition to the function and topic metadata, we annotate each ordinance in \locus{}-v1 with dimensional data.
We consider four dimensions:

\begin{enumerate}
    \item \textbf{Enforcement Discretion} (\textit{highly discretionary} to \textit{non-discretionary}) --- how much selective judgment does the law leave to officials?
    \item \textbf{Opacity} (\textit{opaque} to \textit{intelligible}) --- how hard is it for an ordinary person to know what is required?
    \item \textbf{Paternalism} (\textit{paternalistic} to \textit{externality oriented}) --- is it protecting the actor from themself or protecting others/the public?
    \item \textbf{Problem Salience} (\textit{highly salient} to \textit{unimportant}) ---how strongly does it represent the issue as important, urgent, or threatening?
\end{enumerate}

Examples of laws occupying these dimensions are given in Figure~\ref{fig:axes-demo}.
For instance, the law preventing minors under the age of 16 from attending a festival unless accompanied by an adult is scored as highly paternalistic, intelligible, with neutral discretion and salience.  

Our core intuition is that these dimensions are continuous and that they can better be used to order and measure laws rather than categorize them.
Accurate models of dimensions allow us to bring into focus particular aspects of the law.
Incorporating all laws onto the same set of axes enables analysis both within individual bodies of law (i.e., within a single city), but also for comparative analysis across bodies.

\subsection{Building \locus{} Scorers}
\label{scorers}

For our dimensional analysis, we fine-tune a ModernBERT-base with a linear regression head for each dimension to score a law.
For each dimension, we generate 10,000 scores using 200,000 pairwise LLM-as-a-judge match-ups between ordinances.
During each match, we ask the LLM to compare the two ordinances along a specific dimension, and return which better exemplifies that dimension.
The model outputs A, B, or Tie.
Order can produce bias in pairwise judgement \citep{liu2024aligning}, so every (A, B) comparison pair is also judged in reverse order (B, A).
Pairwise comparison aligns better with human judgement than direct/numeric scoring \citep{liu2024aligning}, motivating us to use it for dimensional scoring.
Each ordinance's match history is used to compute its latent score along each axis using the Bayesian skill rating system, TrueSkill~\citep{herbrich2006trueskill}.
This gives us a total ordering over the sampled ordinances, along with an underlying mean, $\mu$.

To train the regression model, we normalize the scores to their z-score by subtracting out the dimension's mean and dividing its standard deviation. 
For each dimension, we split the 10,000 scored ordinances into a training set (n=8,000), validation set (n=1,000), and test set (n=1,000).
We fine-tune a ModernBERT regression model to predict the normalized TrueSkill score, using mean-squared error as our loss function.
To evaluate the model, we compute Pearson correlation on the test set.
This technique is inspired by the methodology behind Havelock.ai, an AI-powered orality detector that scores text on how oral or literate it is~\citep{havelock_ai}.

\begin{figure}[h!]
    \centering
    \includegraphics[width=\linewidth]{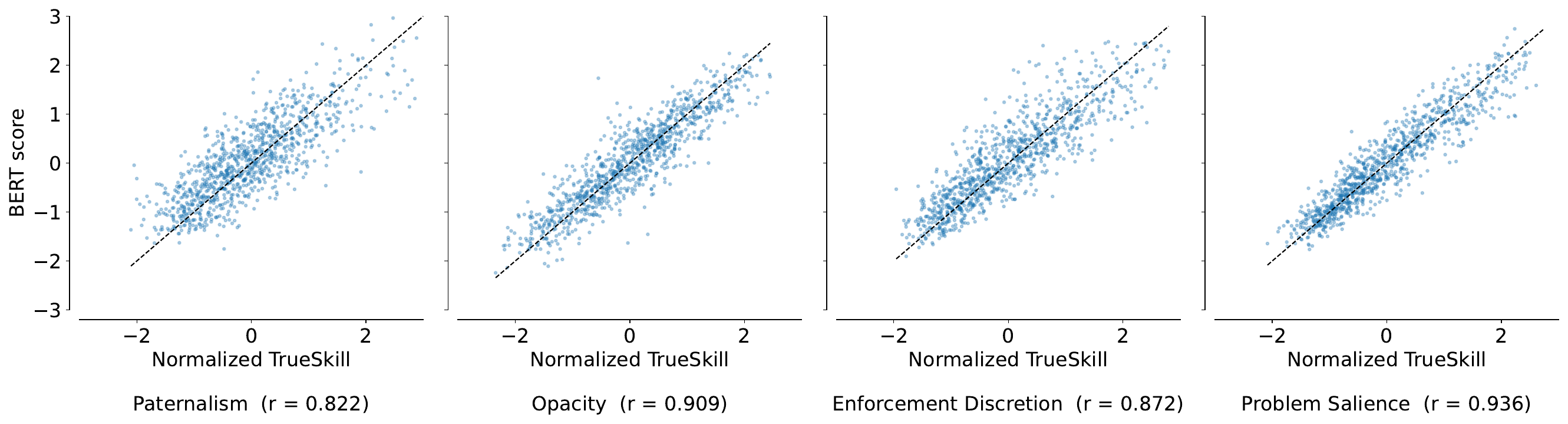}
    \caption{The Pearson correlation between the predicted BERT scores and the normalized TrueSkill scores on 4 distinct test sets (1,000 ordinances per dimension).}
    \label{fig:correlation}
\end{figure}

The dataset for each dimension is constructed using a fixed 10{,}000 ordinance sample, and 200{,}000 pairwise comparisons using GPT-5.4-nano.
We report the Pearson correlation coefficient of the trained BERT models versus the TrueSkill values in Figure~\ref{fig:correlation}.
Each dimension has a correlation of between 0.82 and 0.94, implying the BERT-based scorers largely capture the dynamics of the TrueSkill model.
We provide the prompts plus a sample of high- and low-scoring laws along each dimension in Appendix~\ref{app:prompts}.
We also provide a website to view the TrueSkill scores for the 10{,}000 laws along each dimension.\footnote{\href{https://locallaws--locus-leaderboards-web.modal.run}{https://locallaws--locus-leaderboards-web.modal.run}}
We can use these scores to analyze the laws and correlate them with real-world values of interest, discussed in the next section.

\subsection{Analysis}

Figure~\ref{fig:distribution} demonstrates the importance of studying this at a nationwide rather than a single case level.  For example, counties are notably more opaque than cities on average and   Florida is more than twice as opaque as any other state.  Studying multiple dimensions in tandem can unlock new insights into unique laws; opacity and paternalism are only weakly correlated across sections (Pearson r=0.11 on n=2,211,516).

Finding interesting needles in this haystack of laws can be facilitated through this evaluation. For example, curfews are detected with paternalism and a subsequent analysis of the data provides insight into curfew distribution for minors across the United States.  Headers containing 'possession' and 'alcoholic' are associated with paternalistic laws while 'definitions' and 'variances' are associated with opaque ones.

\section{Discussion, Limitations, and Future Work}
\label{sec:discussion}

LOCUS-v1 is designed as an access layer, not as a final theory of local legal authority. Its county-harmonized release adopts a transparent simplification: for each county, we select the most substantial available local code among the county code and the code of the county's largest municipality. This design makes local law searchable and comparable on a national geographic substrate, but it does not determine which rule controls for a particular person, parcel, business, or legal question. In local law, authority is layered. State statutes, home-rule provisions, county ordinances, municipal codes, charters, preemption doctrines, and issue-specific delegations may all matter. LOCUS therefore should be understood as infrastructure for retrieval, comparison, and benchmark construction rather than as a substitute for doctrine-sensitive legal analysis.

The corpus itself shows why this distinction matters. City and county codes are not interchangeable legal objects. Across the raw corpus, county codes contain substantially more zoning material, while city codes contain more nuisance and public-order regulation. This pattern is consistent with a functional division of local authority: counties more often regulate land, development, and unincorporated territory, while cities more often regulate density, proximity, and everyday public order. For downstream users, this means that jurisdiction type is not merely provenance metadata. It is part of the substantive representation of local law. Models trained or evaluated on local codes should therefore preserve whether a text comes from a municipal or county source, even when the release is harmonized to a county-level unit of analysis.

LOCUS also reveals that local codes share a common representational architecture. When ordinances are ordered by their position in a code, topics tend to appear in a stable sequence: general provisions and governmental structure near the front, followed by business regulation, nuisance and public-order rules, zoning, and building regulation. This finding suggests that local law is not simply a bag of rules. It is organized through a recurring documentary form. That form matters for legal AI. Retrieval systems, chunking strategies, and benchmark designs that ignore position within a code may miss information embedded in the structure of codification itself.

At the same time, LOCUS documents the limits of any simple national harmonization. In much of the country, counties and cities follow the functional pattern described above. In the Northeast, however, the relationship changes: counties appear less zoning-heavy and more enforcement-oriented, consistent with a different institutional history in which towns and municipalities retain more primary land-use authority while counties often perform administrative, health, or enforcement functions. The implication is not that harmonization is impossible. Rather, it is that harmonization must be explicit about what it preserves and what it abstracts away. A county-level substrate is useful because counties form a mutually exclusive and exhaustive national geography, but the legal meaning of a county code is not constant across states and regions.

These limitations point directly to the next generation of legal AI benchmarks. A system that can answer questions about local law must do more than retrieve a plausible ordinance. It must identify the relevant layer of government, distinguish city from county authority, incorporate state-law context, recognize when multiple sources overlap, and reason about whether a retrieved text is actually controlling for the issue at hand. LOCUS-v1 provides the text, metadata, and geographic substrate needed to build those tasks while preserving a clean separation between corpus construction and future evaluations of legal reasoning.

More broadly, LOCUS shows that freeing the law is not only a problem of access. It is a problem of representation. Local ordinances were formally public before LOCUS, but they were not available as a national object of machine reading, systematic comparison, or computational legal analysis. Once made observable at scale, local law appears neither as an undifferentiated mass of rules nor as a set of isolated municipal idiosyncrasies. It has structure: a recurring architecture of codification, a functional division between jurisdictional forms, and regionally specific institutional variation. These are precisely the kinds of structure that legal AI systems must learn to respect if they are to move from text retrieval toward reliable reasoning over public authority.

\medskip

\bibliographystyle{plainnat}
\bibliography{references}

@inproceedings{liu2024aligning,
  title={Aligning with Human Judgement: The Role of Pairwise Preference in Large Language Model Evaluators},
  author={Liu, Yinhong and Zhou, Han and Guo, Zhijiang and Shareghi, Ehsan and Vuli{\'c}, Ivan and Korhonen, Anna and Collier, Nigel},
  booktitle={Conference on Language Modeling (COLM)},
  year={2024}
}

@misc{GeorgetownLawLibrary2026,
  title        = {State Legal Research: General and Multi-Jurisdictional --- Local Government},
  author       = {{Georgetown Law Library}},
  organization = {Georgetown University Law Center},
  howpublished = {\url{https://guides.ll.georgetown.edu/statelegalresearch/localgovernment}},
  note         = {Last updated February 27, 2026; accessed May 5, 2026},
  year         = {2026}
}

@article{johnson2016mimic,
  title={MIMIC-III, a freely accessible critical care database},
  author={Johnson, Alistair E. W. and Pollard, Tom J. and Shen, Lu and Lehman, Li-wei H. and Feng, Mengling and Ghassemi, Mohammad and Moody, Benjamin and Szolovits, Peter and Celi, Leo Anthony and Mark, Roger G.},
  journal={Scientific Data},
  volume={3},
  number={1},
  pages={160035},
  year={2016},
  publisher={Nature Publishing Group},
  doi={10.1038/sdata.2016.35}
}

@article{johnson2023mimic,
  title={MIMIC-IV, a freely accessible electronic health record dataset},
  author={Johnson, Alistair E. W. and Bulgarelli, Lucas and Shen, Lu and Gow, Alvin and Pollard, Tom and Horng, Steven and Celi, Leo Anthony and Mark, Roger},
  journal={Scientific Data},
  volume={10},
  number={1},
  pages={1},
  year={2023},
  publisher={Nature Publishing Group},
  doi={10.1038/s41597-022-01899-x}
}

@inproceedings{guha2023legalbench,
  title={LegalBench: A Collaboratively Built Benchmark for Measuring Legal Reasoning in Large Language Models},
  author={Guha, Neel and Nyarko, Julian and Ho, Daniel E. and R{\'e}, Christopher and Chilton, Adam and Narayana, Aditya and Chohlas-Wood, Alex and Peters, Austin and Waldon, Brandon and Rockmore, Daniel N. and others},
  booktitle={Advances in Neural Information Processing Systems (NeurIPS)},
  year={2023}
}

@inproceedings{chalkidis2022lexglue,
  title={{LexGLUE}: A Benchmark Dataset for Legal Language Understanding in {E}nglish},
  author={Chalkidis, Ilias and Jana, Abhik and Hartung, Dirk and Bommarito, Michael and Androutsopoulos, Ion and Katz, Daniel Martin and Aletras, Nikolaos},
  booktitle={Proceedings of the 60th Annual Meeting of the Association for Computational Linguistics (ACL)},
  year={2022}
}

@inproceedings{henderson2022pile,
  title={Pile of Law: Learning Responsible Data Filtering from the Law and a 256{GB} Open-Source Legal Dataset},
  author={Henderson, Peter and Krass, Mark S. and Zheng, Lucia and Guha, Neel and Manning, Christopher D. and Jurafsky, Dan and Ho, Daniel E.},
  booktitle={Advances in Neural Information Processing Systems (NeurIPS)},
  year={2022}
}

@inproceedings{zheng2021casehold,
  title={When Does Pretraining Help? Assessing Self-Supervised Learning for Law and the {CaseHOLD} Dataset of 53,000+ Legal Holdings},
  author={Zheng, Lucia and Guha, Neel and Anderson, Brandon R. and Henderson, Peter and Ho, Daniel E.},
  booktitle={Proceedings of the 18th International Conference on Artificial Intelligence and Law (ICAIL)},
  year={2021}
}

@inproceedings{hendrycks2021cuad,
  title={{CUAD}: An Expert-Annotated {NLP} Dataset for Legal Contract Review},
  author={Hendrycks, Dan and Burns, Collin and Chen, Anya and Ball, Spencer},
  booktitle={Advances in Neural Information Processing Systems (NeurIPS), Datasets and Benchmarks Track},
  year={2021}
}

@inproceedings{koreeda2021contractnli,
  title={{ContractNLI}: A Dataset for Document-Level Natural Language Inference for Contracts},
  author={Koreeda, Yuta and Manning, Christopher D.},
  booktitle={Findings of the Association for Computational Linguistics: EMNLP},
  year={2021}
}

@inproceedings{holzenberger2020sara,
  title={A Dataset for Statutory Reasoning in Tax Law Entailment and Question Answering},
  author={Holzenberger, Nils and Blair-Stanek, Andrew and Van Durme, Benjamin},
  booktitle={Proceedings of the Natural Legal Language Processing Workshop},
  year={2020}
}

@article{dahl2024hallucinating,
  title={Large Legal Fictions: Profiling Legal Hallucinations in Large Language Models},
  author={Dahl, Matthew and Magesh, Varun and Suzgun, Mirac and Ho, Daniel E.},
  journal={Journal of Legal Analysis},
  volume={16},
  number={1},
  pages={64--93},
  year={2024}
}

@inproceedings{tuggener2020ledgar,
  title={{LEDGAR}: A Large-Scale Multi-Label Corpus for Text Classification of Legal Provisions in Contracts},
  author={Tuggener, Don and von D{\"a}niken, Pius and Peetz, Thomas and Cieliebak, Mark},
  booktitle={Proceedings of the 12th Language Resources and Evaluation Conference (LREC)},
  year={2020}
}

@article{herbrich2006trueskill,
  title={TrueSkill™: a Bayesian skill rating system},
  author={Herbrich, Ralf and Minka, Tom and Graepel, Thore},
  journal={Advances in neural information processing systems},
  volume={19},
  year={2006}
}

@article{taghadouini2026lightonocr,
  title={LightOnOCR: A 1B End-to-End Multilingual Vision-Language Model for State-of-the-Art OCR},
  author={Taghadouini, Said and Cavaill{\`e}s, Adrien and Aubertin, Baptiste},
  journal={arXiv preprint arXiv:2601.14251},
  year={2026}
}

@article{bai2025qwen3,
  title={Qwen3-vl technical report},
  author={Bai, Shuai and Cai, Yuxuan and Chen, Ruizhe and Chen, Keqin and Chen, Xionghui and Cheng, Zesen and Deng, Lianghao and Ding, Wei and Gao, Chang and Ge, Chunjiang and others},
  journal={arXiv preprint arXiv:2511.21631},
  year={2025}
}

@article{poznanski2025olmocr,
  title={olmocr 2: Unit test rewards for document ocr},
  author={Poznanski, Jake and Soldaini, Luca and Lo, Kyle},
  journal={arXiv preprint arXiv:2510.19817},
  year={2025}
}

@inproceedings{warner2025smarter,
  title={Smarter, better, faster, longer: A modern bidirectional encoder for fast, memory efficient, and long context finetuning and inference},
  author={Warner, Benjamin and Chaffin, Antoine and Clavi{\'e}, Benjamin and Weller, Orion and Hallstr{\"o}m, Oskar and Taghadouini, Said and Gallagher, Alexis and Biswas, Raja and Ladhak, Faisal and Aarsen, Tom and others},
  booktitle={Proceedings of the 63rd Annual Meeting of the Association for Computational Linguistics (Volume 1: Long Papers)},
  pages={2526--2547},
  year={2025}
}

@misc{havelock_ai,
  title        = {Havelock AI},
  author       = {Weisenthal, Joe},
  howpublished = {\url{https://havelock.ai}},
  year         = {2026},
  note         = {Accessed: 2026-05-06}
}

@misc{openai2026gpt54,
  title        = {Introducing {GPT-5.4}},
  author       = {{OpenAI}},
  year         = {2026},
  month        = mar,
  day          = {5},
  howpublished = {\url{https://openai.com/index/introducing-gpt-5-4/}},
  note         = {Accessed: 2026-05-07}
}

@inproceedings{zheng2023judging,
  title     = {Judging {LLM-as-a-Judge} with {MT-Bench} and {Chatbot Arena}},
  author    = {Zheng, Lianmin and Chiang, Wei-Lin and Sheng, Ying and Zhuang, Siyuan and Wu, Zhanghao and Zhuang, Yonghao and Lin, Zi and Li, Zhuohan and Li, Dacheng and Xing, Eric P. and Zhang, Hao and Gonzalez, Joseph E. and Stoica, Ion},
  booktitle = {Advances in Neural Information Processing Systems},
  volume    = {36},
  year      = {2023},
  eprint    = {2306.05685},
  archivePrefix = {arXiv},
  primaryClass  = {cs.CL}
}

@misc{georgia_v_public_resource_2020,
  title        = {Georgia v. Public.Resource.Org, Inc.},
  author       = {{Supreme Court of the United States}},
  year         = {2020},
  month        = apr,
  day          = {27},
  note         = {590 U.S. 255, 140 S. Ct. 1498, 206 L. Ed. 2d 732},
  howpublished = {Slip Opinion No. 18-1150},
  url          = {https://www.supremecourt.gov/opinions/19pdf/18-1150_7m58.pdf}
}

@article{holmes1897path,
  author  = {Holmes, Jr., Oliver Wendell},
  title   = {The Path of the Law},
  journal = {Harvard Law Review},
  volume  = {10},
  number  = {8},
  pages   = {457--478},
  year    = {1897},
  month   = mar,
}

@misc{caselaw_access_project,
  author       = {{Harvard Law School Library Innovation Lab}},
  title        = {Caselaw {A}ccess {P}roject},
  year         = {2018},
  howpublished = {\url{https://case.law/}},
}

@article{livermore2017supreme,
  author  = {Livermore, Michael A. and Riddell, Allen B. and Rockmore, Daniel N.},
  title   = {The Supreme Court and the Judicial Genre},
  journal = {Arizona Law Review},
  volume  = {59},
  pages   = {837--901},
  year    = {2017}
}

@article{aletras2016predicting,
  author  = {Aletras, Nikolaos and Tsarapatsanis, Dimitrios and
             Preo{\c{t}}iuc-Pietro, Daniel and Lampos, Vasileios},
  title   = {Predicting Judicial Decisions of the {E}uropean {C}ourt of
             {H}uman {R}ights: A Natural Language Processing Perspective},
  journal = {PeerJ Computer Science},
  volume  = {2},
  pages   = {e93},
  year    = {2016},
  doi     = {10.7717/peerj-cs.93}
}

@inproceedings{steinberger2006jrc,
  author    = {Steinberger, Ralf and Pouliquen, Bruno and Widiger, Anna and
               Ignat, Camelia and Erjavec, Toma{\v{z}} and Tufi{\c{s}}, Dan and
               Varga, D{\'a}niel},
  title     = {The {JRC-Acquis}: {A} Multilingual Aligned Parallel Corpus
               with 20+ Languages},
  booktitle = {Proceedings of the 5th International Conference on
               Language Resources and Evaluation (LREC)},
  year      = {2006}
}

@inproceedings{chalkidis2019neural,
  author    = {Chalkidis, Ilias and Androutsopoulos, Ion and Aletras, Nikolaos},
  title     = {Neural Legal Judgment Prediction in {E}nglish},
  booktitle = {Proceedings of the 57th Annual Meeting of the
               Association for Computational Linguistics (ACL)},
  pages     = {4317--4323},
  year      = {2019}
}

\appendix

\section{Scoring Prompts}
\label{app:prompts}

We elicit pairwise judgments from GPT-5.4-nano using a single shared
template, parameterized by a rubric for each axis.

\begin{promptbox}{Pairwise Comparison System Prompt}
You are evaluating local laws along the dimension of "{{ axis }}"
{%
Read the following two laws and determine which has a GREATER degree of {{ axis }} according to the rubric above.
Respond with the winner and a one sentence explanation of why it is the winner.
-----
Law A
-----
{{ law_a["header"] }}
{{ text_a }}
-----
Law B
-----
{{ law_b["header"] }}
{{ text_b }}
-----
Respond in the following JSON format only:
```json
{
    "winner": "A" or "B" or "Tie"
    "reasoning": "one sentence explanation"
}
```
\end{promptbox}

\begin{promptbox}{Axis Rubric: Problem Salience}
# Problem Salience
Problem salience measures how strongly the law represents the regulated issue as important, urgent, or threatening.
A high-salience law uses charged framing (crisis, epidemic, threat), findings/preambles emphasizing severity, or heightened penalties signaling gravity.
A low-salience law treats the issue as routine, technical, or administrative without rhetorical emphasis on stakes.
\end{promptbox}

\begin{promptbox}{Axis Rubric: Paternalism vs.\ Externality Orientation}
# Paternalism vs. Externality Orientation
Paternalism vs. externality orientation measures whether the law is primarily protecting the regulated actor from themself or protecting others/the public from the actor's conduct.
A highly paternalistic law targets self-regarding behavior (harms or risks borne mainly by the actor).
A law oriented toward externalities targets conduct whose harms fall on third parties or the public at large.
\end{promptbox}

\begin{promptbox}{Axis Rubric: Opacity / Intelligibility}
# Opacity / Intelligibility
Opacity / intelligibility measures how hard it is for an ordinary person to know what the law requires of them. A highly opaque law relies on dense cross-references, technical jargon, undefined terms, or convoluted structure that obscure the obligations.
A low-opacity (highly intelligible) law states its requirements in plain, self-contained language a layperson can readily understand.
\end{promptbox}

\begin{promptbox}{Axis Rubric: Enforcement Discretion}
# Enforcement Discretion
Enforcement discretion measures the degree to which a citizen's exposure to enforcement under a law depends on official choice rather than on the citizen's own conduct.
It is high when two factors compound: (1) breadth of exposure -- the pool of citizens potentially subject to the law is large because its triggering conditions are vague, evaluative, or so commonly met that many qualify as eligible targets; and (2) textual latitude -- the statute's language gives officials wide freedom over whether, when, against whom, and how to act.
A law that exposes a vast pool but mandates uniform enforcement leaves officials little real choice; a law that grants officials sweeping latitude but exposes no citizens to enforcement at all -- e.g., provisions concerning only internal government structure, personnel, contracting, or interagency procedure -- creates no opportunity for capricious wielding.
The score floor is reserved for laws1 that do not act on private parties; the score ceiling for laws under which many citizens stand exposed and officials choose freely among them.
\end{promptbox}

\section{Annotation Prompt}
We prompt gpt-5.4-nano for an initial zero-shot classification, and review anything evaluated flagged annotations (5.5\%) with a second pass of gpt-5.4.  

\begin{promptbox}{Annotation Prompt}
SYSTEM_PROMPT = (
    "You are a legal text classifier specializing in municipal and county "
    "codes. Return only a JSON object that matches the provided schema."
)

REVIEW_SYSTEM_PROMPT = (
    "You are a senior legal QA reviewer. Review the first-pass classification "
    "carefully, correct it when needed, and return only a JSON object that "
    "matches the provided schema. Treat Process as the label for operative "
    "administrative procedures, delegated authority, internal governance rules, "
    "meeting/quorum/voting rules, board composition, appointments, elections, "
    "hearings, notice requirements, and permitting workflows. Use Structural "
    "only for non-operative artifacts or formatting noise such as headers, "
    "tables of contents, HTML fragments, history/source notes, page markers, "
    "cross-reference lists, and similar text that does not itself state an "
    "operative rule or procedure."
)

USER_INSTRUCTIONS = """
Task: Classify the provided text by its primary legal function.

Allowed primary_function values:
- Context: defines terms, scope, or introductory intent.
- Rules: imposes permissions, obligations, or prohibitions.
- Process: describes administrative procedure, authority, or government structure.
- Enforcement: specifies penalties, violations, appeals, or exceptions.
- Structural: non-substantive artifacts such as section headers, tables of
  contents, HTML, history/source notes, page numbers, or formatting remnants.

Output rules:
- is_substantive must be 1 only for Rules or Enforcement. Otherwise use 0.
- primary_function must be exactly one of: Context, Rules, Process,
  Enforcement, Structural.
- sub_category must be null when is_substantive is 0.
- When is_substantive is 1, sub_category must be exactly one of:
  Land use, Noise/Nuisance, Housing, Business licensing, Public space,
  Building/Safety, Other.
- logic must be one short sentence.
""".strip()

REVIEW_INSTRUCTIONS = """
Task: Review a first-pass legal text classification and produce the final
corrected classification.

Rules:
- If the first-pass classification is correct, keep it and set review_outcome to
  "confirm".
- If the first-pass classification is incorrect, correct it and set
  review_outcome to "override".
- If the first-pass classification is missing or invalid, classify from scratch
  and set review_outcome to "fresh".
- Keep the same classification rules and category set as the first pass.
- Treat operative internal governance text as Process, not Structural. This
  includes board composition, quorum, voting, meeting procedures, delegation of
  authority, appointment/removal rules, election administration, hearings,
  notice, application steps, and similar administrative workflows.
- Use Structural only for non-operative artifacts/noise such as section headers,
  article labels, tables of contents, page markers, history/source notes, HTML,
  formatting remnants, or cross-reference lists that do not themselves contain
  operative requirements.
- Derive is_substantive from primary_function: use 1 only for Rules or
  Enforcement; use 0 for Context, Process, and Structural.
- review_logic must briefly explain why you confirmed, changed, or freshly
  classified the text.
""".strip()

CLASSIFICATION_SCHEMA = {
    "type": "object",
    "additionalProperties": False,
    "properties": {
        "is_substantive": {"type": "integer", "enum": [0, 1]},
        "primary_function": {
            "type": "string",
            "enum": [
                "Context",
                "Rules",
                "Process",
                "Enforcement",
                "Structural",
            ],
        },
        "sub_category": {
            "anyOf": [
                {
                    "type": "string",
                    "enum": [
                        "Land use",
                        "Noise/Nuisance",
                        "Housing",
                        "Business licensing",
                        "Public space",
                        "Building/Safety",
                        "Other",
                    ],
                },
                {"type": "null"},
            ]
        },
        "logic": {"type": "string"},
    },
    "required": [
        "is_substantive",
        "primary_function",
        "sub_category",
        "logic",
    ],
}
\end{promptbox}

\newpage

\end{document}